\documentclass[runningheads]{llncs}
\usepackage[T1]{fontenc}
\usepackage{graphicx}
\usepackage{booktabs}
\usepackage[misc]{ifsym}
\newcommand{\corr}{(\Letter)}
\usepackage{booktabs}
\usepackage{makecell,multirow}
\usepackage{amsmath,amsfonts,bm}

\def\1{\bm{1}}

\DeclareMathAlphabet{\mathsfit}{\encodingdefault}{\sfdefault}{m}{sl}
\SetMathAlphabet{\mathsfit}{bold}{\encodingdefault}{\sfdefault}{bx}{n}

\usepackage{xcolor}
\usepackage{import}
\usepackage{pifont}
\usepackage{subcaption} % For subfigures
\usepackage{array} % For easier table formatting

\newcolumntype{H}{>{\setbox0=\hbox\bgroup}c<{\egroup}@{}} % Handy to hide columns

\usepackage[textsize=tiny]{todonotes}

\usepackage{url}
\usepackage{breakcites}
\usepackage[colorlinks=true,breaklinks=true]{hyperref}
\begin{document}

\title{Early Yield Prediction for Sugar Beet Fields using Satellite Data - Learnings from Specialized Vision Transformers}

\titlerunning{Early Sugar Beet Yield Prediction using Satellite Data}

\author{
Philipp Vaeth\inst{1,2}\corr
\and \\
Bhumika Laxman Sadbhave\inst{1}
\and \\
Denise Dejon \inst{3}
\and \\
Gunther Schorcht \inst{3}
\and \\
Magda Gregorová\inst{1}}

\authorrunning{Vaeth et al.}
\institute{Center for Artificial Intelligence and Robotics, Technical University of Applied Sciences Wuerzburg-Schweinfurt, Franz-Horn-Strasse 2,
Wuerzburg, Germany\\
\email{\{philipp.vaeth,magda.gregorova\}@thws.de} \and 
Bielefeld University, Universitaetsstrasse 25, Bielefeld, Germany\\ \and
green spin GmbH,  Magdalene-Schoch-Strasse 5, Wuerzburg, Germany}

\maketitle              % typeset the header of the contribution

\begin{abstract}
Remote sensing has become an increasingly valuable tool for agricultural monitoring, particularly through the use of publicly available satellite imagery. 
However, effectively integrating domain knowledge into machine learning methods remains challenging.
This study presents a real-world example of early sugar beet harvest yield forecasting from purely optical Sentinel-2 imagery, demonstrating how a tight integration of domain knowledge and machine learning can lead to synergistic gains.
We empirically find that using very small vision transformer patch sizes and all available Sentinel-2 spectral bands improves our model despite being uncommon design choices in the domain.
As a practical contribution, we were able to identify a large fraction of low-yield fields in a different year early on in the growth cycle through a modified training setup and a ranking-based detection of underperforming fields.

\keywords{Remote sensing  \and Sentinel-2 \and Satellite \and Vision Transformer \and ViT \and Yield Prediction.}
\end{abstract}

\section{Introduction}
Remote sensing, especially satellite imagery, has gained interest from the machine learning community in recent years~\cite{survey}.
The scalability of modern machine learning methods, in combination with the large volume of satellite imagery, has led to improvements in many areas, for example in optical crop yield estimation~\cite{xiao2025progress,agriculture16040417}.
The domain presents several technical challenges, including incomplete time series, varying object sizes, high-dimensional inputs, measurement errors, extremely large data volumes, and scarce labels.
These challenges require non-trivial task-specific adaptations of standard computer vision models.
Unfortunately, such adaptations are rarely transferable to other tasks, because different applications may require fundamentally different receptive fields, time resolutions, and spectral bands.

This study focuses on sugar beet yield prediction purely from optical satellite imagery, a prominent, inexpensive, and scalable resource in remote sensing that is nevertheless rarely used as the sole source of information.
We gather a data set of several thousand sugar beet field locations in Bavaria, Germany, for the years 2019 and 2024, the corresponding yield in tons per hectare (t/ha) at the end of the harvest, and the publicly available Sentinel-2 satellite images of the fields throughout the growth cycle.
The sugar beet (beta vulgaris) is an important crop in Europe for various industries such as livestock, fermentation, biofuel and export.
The amount of sugar in the harvest depends largely on location, weather conditions, as well as prevalence of diseases or other stress factors that occur during the growth cycle~\cite{sadbhave2025sugarbeetstressdetectionusing}.

Many existing deep-learning yield predictors rely on multi-source data such as climate data~\cite{kaur2023fusion,liu2023attention,schwalbert2020satellite}, multi-spectral information from other sources~\cite{lu2024deep}, data from Unmanned Aerial Vehicle (UAVs)~\cite{YANG2021106092,maimaitijiang2020soybean}, soil data~\cite{kaur2023fusion} or fine-grained labels~\cite{jeong2022predicting}.
The models also use task-specific combinations of spectral bands~\cite{liu2023attention,schwalbert2020satellite,lu2024deep,maimaitijiang2020soybean,jeong2022predicting} and custom receptive fields~\cite{liu2023attention,lu2024deep,YANG2021106092,jeong2022predicting} for their models. 
These modeling prerequisites make their design choices and results hard to transfer to other yield prediction tasks.
Instead of providing a new competitive prediction system, we isolate the yield prediction performance to a purely optical, single-label-per-field setting and empirically validate critical design choices.
We make the following contributions:
\begin{itemize}
      \item We formulate the sugar beet yield prediction task based on purely optical Sentinel-2 satellite data as a supervised regression problem at the field level across several thousand fields in Bavaria and a held-out test year motivated by the practical application.
      \item We systematically study ViT design choices (patch size, number of spectral bands, positional encodings, embedding size and heads) for this setting and show that very small patches and using all Sentinel-2 bands are critical for performance.
      \item We extend the model with temporal encodings to enable early prediction and a ranking-based detection of underperforming fields. Our results show that, even though the absolute yield prediction is extremely challenging on purely optical data, a large fraction of low-yield fields can be detected well before the harvest.
\end{itemize}

\section{Methods}\label{sec:methods}
In this section, we describe the dataset, the prediction task, pre-processing steps, the model setup, training details, the experimental setup, the evaluation, and the limitations.
\paragraph{Data set}
This paper uses publicly available Sentinel-2 Level 2A satellite images~\cite{sentinel2}.
The Sentinel-2 satellite takes an image every 5 days (called temporal instances) with bands of different spatial resolutions. 
Four bands at 10 meter resolution contain the classic RGB values and a "near-infrared" (NIR) band.
Six bands at 20 meter resolution contain four "visible and near-infrared" (VNIR) bands and two "short-wave infrared" (SWIR) bands for applications such as snow, ice, cloud detection or vegetation moisture stress assessment.
Three bands at 60 meter spatial resolution focus on cloud screening, atmospheric correction and cirrus detection.
The 20 m bands are resampled to 10 m spatial resolution for input consistency.
The three 60 m bands are discarded due to redundancy with a cloud mask described in the following paragraph.
For reference, a patch of 16 pixels in the data at a ground resolution of 10 m corresponds to a real-world area of over three and a half soccer fields ($25600\ m^2$).

\paragraph{Task definition}
In addition to the publicly available satellite images, we have a cloud mask, a field mask and the corresponding harvest yield available in the project.
This custom data set contains 2614 fields from the year 2019 and 5695 fields from the year 2024.
The goal is to predict the field-level sugar yield based on the last Sentinel-2 satellite image before the harvest.
We purposefully do not use the temporal nature of the satellite data until later in the paper (section~\ref{sec:model_application}), in order to empirically test design choices when using only the optical information in the images.
This means we formulate the task as an image regression problem with only spatial and channel information from the satellite just before the harvest.
We are unable to disclose the data set of the project at the current time due to contractual constraints with the industrial partner and privacy of farm locations, but we will explain our pre-processing and modeling choices as clearly as possible.

\begin{figure}[t]
      \centering
      \includegraphics[width=0.9\linewidth]{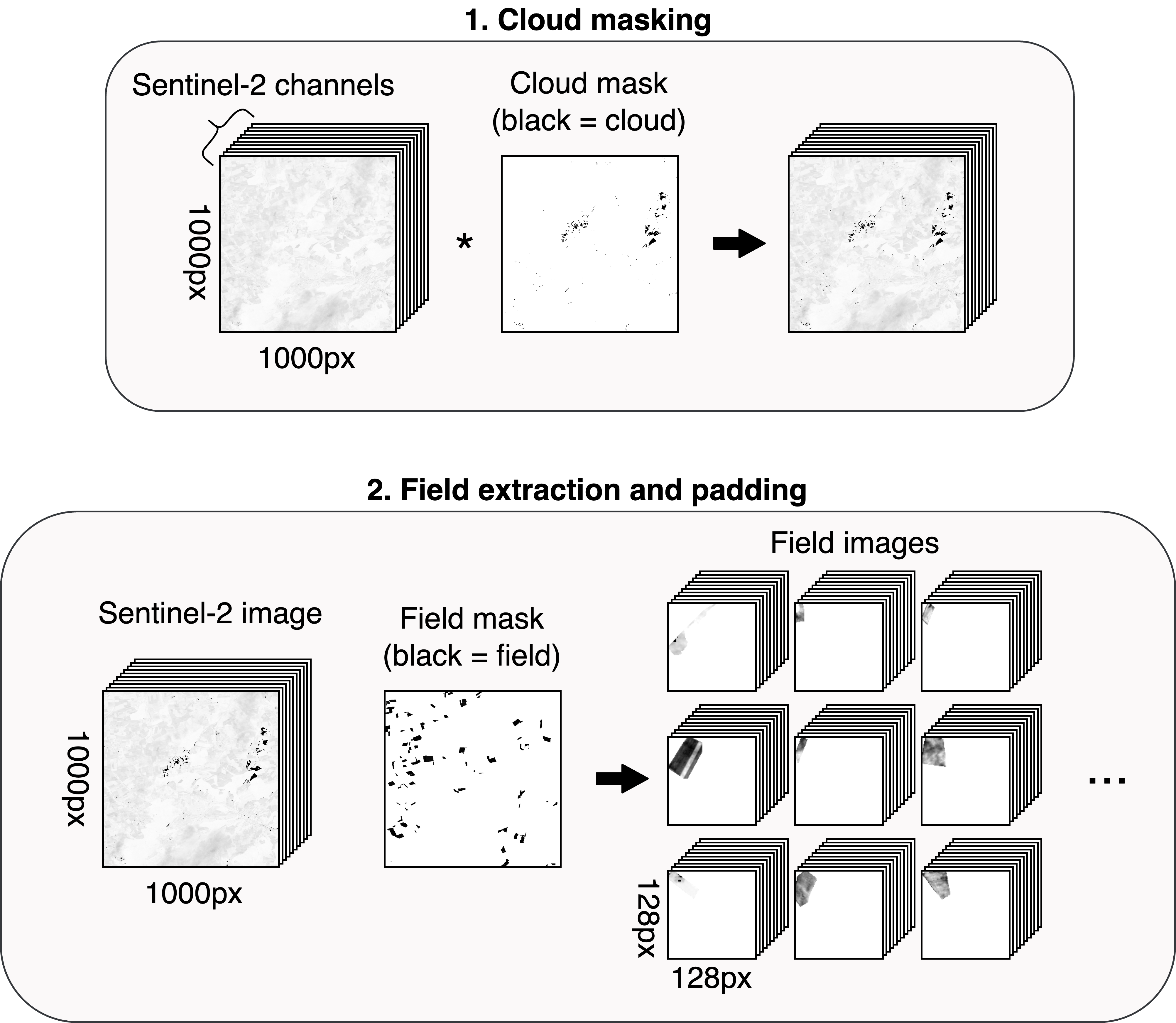}
      \caption{Data pre-processing pipeline}\label{fig:preprocessing}
\end{figure}
\paragraph{Pre-processing} 
The Sentinel-2 data is provided as chunks of $1000 \times1000$~px images with 10 channels.
To be able to use machine learning techniques on this data, we need to first pre-process the images.
We deliberately chose to keep the pre-processing steps minimal and rather rely on the end-to-end training of our model to extract information.
We provide a full overview of the pre-processing steps in figure~\ref{fig:preprocessing}.
First, we filter defective measurements through a cloud mask.
Afterwards, we extract each field into a separate field image.
The locations and shapes of the fields are given by the field mask.
Due to inconsistent field sizes, the pixels are placed on the top left and padded to the model input size of $128 \times 128$~px, such that there is always only one complete field per field image and shapes of fields are preserved.
We chose this input size to fit the largest field in the data set, which however leads to additional compute overhead for fields that are much smaller.
To prevent the model from inferring implicit information from the surrounding context (such as the field location), we remove non-field pixels during the padding operation.
For the empirical study in section~\ref{sec:experiments}, we select the last temporal instance before the harvest for each field.
Later in section~\ref{sec:model_application}, we extend the model to train on all temporal instances through time encodings to enable early prediction.
We do not condition the model on other information such as acquisition dates, longitude or latitude in order to prevent the model from overly relying on this conditioning information rather than purely on the optical data.

\paragraph{Model setup \& training}
We mainly consider deep neural networks for the end-to-end feature extraction but also include two non-deep baselines in our experiments.
The simplest baseline is predicting the mean yield of the fields in the training set.
For a simple domain-specific baseline, we train a linear regression model on statistics (min, mean, max) of the NDVI, EVI, MSI vegetation indices (see section~\ref{sec:channels}).
Convolution-based models are directly applied with only minor modifications in the first layer to accommodate for the 10 Sentinel-2 channels and treat them exactly as in a three channel RGB setup.
The Vision Transformer~\cite{vit} operates over a sequence of non-overlapping patches of a field image.
Inspired by the class embedding from the original ViT, we add an extra token to aggregate information from all patches during the attention operation into a single embedding.
We also considered predicting a harvest yield per patch and aggregating this information across all patches in a field by simply adding the values, but this would lead to a strong model assumption that all patches are similarly important as the gradients in the backwards pass would be equally distributed across all patches.
Preliminary experiments showed better performance through the learned token aggregation.

After the feature extraction by the models, a single linear layer predicts the harvest yield per field image based on the final feature vector.
For faster training convergence, the bias of the regression layer is initialized to the mean yield of the training set.
We intentionally kept this regression layer simple to limit the influence of the classifier and to enable clear design conclusions about the end-to-end feature extractor.
The 2019 data will be split into training and validation sets (70:30), and the 2024 data will be the test set for the evaluation of the model.
We train all models with an RMSE objective on the harvest yield, a learning rate of 0.001, a batch size of 128, a maximum of 150 epochs and include early stopping based on a running average of the validation RMSE loss.

\paragraph{Experimental setup \& evaluation} 
For our experiments, we initially predict the harvest yield from the last satellite image before the harvest (section~\ref{sec:experiments}) and then extend the model to early prediction based on images from early on in the growth cycle (section~\ref{sec:model_application}).
We measure the performance of the harvest yield prediction through an RMSE objective as a standard metric in the machine learning community.
We train the model on the 2019 training data and report the 2019 validation RMSE for different models and hyperparameter configurations.
We use the 2024 data as a hold-out test set to align with the practical application, as models are trained on historical data to predict the yield for new satellite images of an upcoming season.
In our setup, this also serves as an out-of-distribution test to ensure the learned optical features are truly capturing yield-specific features, as the 2019 and 2024 seasons had fundamentally different climate and growth conditions.
For the early prediction in section~\ref{sec:model_application}, we introduce a ranking-based metric to identify underperforming fields.
We compare the complexity of all baselines and model design choices by reporting the parameter count and an estimate of compute requirements for a single forward pass (in GFLOPS).
We show a pareto-style plot for the performance versus complexity trade-off for all experiments in figure~\ref{fig:pareto}.

\paragraph{Limitations}
As a result of our purely optical-based data set and the task-driven experimental setup across two years, comparison with previous work is challenging.
Adapting previous work to our data set would not lead to a fair comparison, as methods require specific choices of spectral bands and task-dependent receptive fields (for example~\cite{xie2021integration,wang2020winter,xiao2024winter,mathivanan2022utilizing,lu2024deep,li2022improving}).
Previous methods are also usually built on additional sources of information and more conditioning variables such as longitude, latitude, acquisition dates, climate data~\cite{SONG2024110101,kaur2023fusion}, multi-spectral information from other sources~\cite{lu2024deep}, lidar data~\cite{LI2024103643}, soil data~\cite{kaur2023fusion}, or additional crop growth models~\cite{chen2024coupling,zhuo2023improved}.
We also considered using recent foundational models (such as~\cite{jakubik2025terramindlargescalegenerativemultimodality,simeoni2025dinov3,satmae,hong2024spectralgpt,brown2025alphaearthfoundationsembeddingfield}) as pre-trained feature extractors, but adapting them to our prediction task with our limited size data set would require substantial model modifications and is beyond the scope of this study.
We instead aim to provide rigorous experiments starting from multiple simpler baseline models that are all adapted to our specific task and then further modified by testing critical design choices that may help future work to develop specialized models for remote sensing tasks.

\section{Experiments}\label{sec:experiments}
In this section, we empirically test important design choices in a hypothesis driven way.
We compare simple baseline model architectures in section~\ref{sec:baselines}, experimentally test the importance of the receptive field in section~\ref{sec:patch_size}, assess the relevance of Sentinel-2 bands in section~\ref{sec:channels}, question whether positional encodings are beneficial for the task in section~\ref{sec:positional_encodings}, and show the complexity versus performance trade-off in section~\ref{sec:dim_heads}.
We conclude this section by summarizing our findings and putting them into broader context in section~\ref{sec:summary}.

\subsection{Which baseline architecture is most suitable for the task?}\label{sec:baselines}
Recent yield prediction approaches use deep learning to leverage the large amount of data in the remote sensing field for more accurate forecasting~\cite{wang2020winter,tao2021corn,li2022improving,xie2021integration,xiao2024winter,jiang2020deep,mathivanan2022utilizing}.
Methods are commonly tailored to the task and auxiliary data available in their projects, making comparisons challenging.
Especially convolution-based neural networks are popular in the domain, whereas newer transformer architectures are becoming more popular in recent years~\cite{joshi2023remote}.

We start our experimental analysis with the simplest baseline, which predicts the mean yield of the fields in the training set.
For a domain-specific baseline similar to hand-crafted feature engineering approaches~\cite{xiao2025progress}, we train a linear regression model on min, mean and max values of the NDVI, EVI and MSI vegetation indices with negligible model parameters and compute requirements.
We consider standard convolution-based models for our sugar beet yield prediction including SqueezeNet~\cite{SqueezeNet}, ResNet-18 and ResNet-50~\cite{Resnet}.
SqueezeNet was chosen as a light-weight convolution-based model.
We also include a single-layer vision transformer (ViT~\cite{vit}) with one 64-dimensional head and a patch size of eight.
The hyperparameter choices for the ViT model were inspired by standard configurations for natural images and by preliminary experiments.
Although there are no direct comparisons to previous methods, we provide a comparison of widely used image backbones with varying capacities and simple, non-deep baselines.

\begin{table}[t]
\centering
\caption{The performance and computational requirements for various baseline models on the sugar beet yield prediction task.}
\label{tab:baselines}
\begin{tabular}{@{}ccccc@{}}
\toprule
      & \thead{Validation \\ RMSE} & \thead{Test \\ RMSE} & \thead{Parameters \\ (million)} & GFLOPS \\ \cmidrule(lr){2-5}
Train set mean &   18.50                     &  22.06         &      -               & -         \\ 
Linear Regression & 18.49 & 22.06 & 0 & 0 \\
ResNet-50 &     17.96                 &    22.30       &     23.66       & 366                  \\ 
ResNet-18 &    18.08                    &   20.73        &     11.23      & 175                \\ 
SqueezeNet & 17.59  & 20.70 & 0.759 & 25 \\ 
ViT baseline &       \textbf{17.50}                &    \textbf{20.48}       &  0.322          & 21                   \\ \bottomrule
\end{tabular}
\end{table}

The results in table~\ref{tab:baselines} show that, despite being the smallest model both in terms of trainable parameters and floating point operations, the ViT baseline outperforms all other models.
We hypothesize this is due to different inductive priors in the architectures that can be exploited in combination with domain knowledge. 
The ViT allocates most compute on the attention mechanism between patches of the input as a fixed receptive field to learn differences between parts of the field. 
The CNN counterparts spread out compute over an increasing receptive field through network depth, while compressing spatial information with the assumption that close pixels are more related.
For our specific yield prediction task where each pixel already represents many plants, we hypothesize that aggregating information over an increasing receptive field is likely not as important as for natural RGB images.

\subsection{How important is the receptive field for the model quality?}\label{sec:patch_size}
It is common in previous work to adapt the models to the task at hand, specifically the receptive field of the model through hyperparameter configurations~\cite{xie2021integration,wang2020winter,xiao2024winter,mathivanan2022utilizing,lu2024deep,li2022improving}.
We believe this is an important design consideration and should be motivated by domain knowledge.
The patch size of the vision transformer in our case defines the receptive field for the model and is a major factor for compute requirements.
A larger patch size leads to fewer patches for the attention mechanism and a higher compression rate as more input pixels are represented by a fixed dimensional vector.
For natural RGB images, the higher compression rate is usually not critical, as objects used for downstream tasks are made up of multiple pixels.
In our case, we hypothesize that the patch size is much more important as each pixel already represents a large part of a field with extremely many plants (1~pixel = 100~$m^2$), where small differences between areas of the fields are important indicators for the harvest yield.
The recommended ViT patch size for natural images is $p=16$~\cite{vit} and many remote sensing foundational models choose similar patch sizes for the Sentinel-2 data~\cite{satmae,jakubik2025terramindlargescalegenerativemultimodality,simeoni2025dinov3}.
We show the effect of the vision transformer patch size on the performance of our specific harvest yield prediction in table~\ref{tab:vit_patch_size}.
We consider the patch sizes \(p\in[1, 2, 4, 8, 16, 32]\) with the common notation of ViT/$p$.
\begin{table}[t]
\centering
\caption{A comparison of the simple vision transformer across different patch sizes for the sugar beet yield prediction task.}
\label{tab:vit_patch_size}
\begin{tabular}{@{}ccccc@{}}
\toprule
      & \thead{Validation \\ RMSE} & \thead{Test \\ RMSE} &  \thead{Parameters \\ (million)} & GFLOPS \\ \cmidrule(lr){2-5}
      ViT/32 &       18.20                 &    21.64       &  0.937              & 3               \\ 
      ViT/16 &       18.00                 &    21.07       &  0.445              & 7               \\ 
      ViT/8 &   17.50                     &  20.48         &      0.322          & 21            \\ 
      ViT/4 &   17.45                    &  20.64         &     0.292         & 75               \\ 
      ViT/2 &   \textbf{17.12}                    &  \textbf{20.41}         &      0.284        & 295                \\ 
      ViT/1 &   17.65                     &  20.81         &      0.282      & 1172                  \\ \bottomrule
\end{tabular}
\end{table}

The results in table~\ref{tab:vit_patch_size} confirm our hypothesis that smaller patch sizes are critical for performance.
The model error on both the validation and the test set generally decreases with a lower patch size, whereas the compute requirements increase due to a higher amount of patches for the attention mechanism. 
The best model has a patch size of $p=2$.
Interestingly, the model with a patch size of $p=1$, which results in self-attention on a pixel level, shows worse performance than some models operating over patched inputs.
We hypothesize this is due to the high amount of compute required to train the pixel-level model, which would require longer training times, a larger data set and different hyperparameters. 
We will continue our experiments with the best ViT/2 model and leave the limitations of the pixel-level transformer to future work.

\subsection{Are all Sentinel-2 input bands necessary for a good model?}\label{sec:channels}
It is common in previous work to use task-specific combinations of spectral bands~\cite{liu2023attention,schwalbert2020satellite,lu2024deep,maimaitijiang2020soybean,jeong2022predicting}.
Common combinations include the Normalized Difference Vegetation Index ($\text{NDVI}=\frac{\text{NIR}-\text{RED}}{\text{NIR}+\text{RED}}$), the MODIS Enhanced Vegetation Index ($\mathrm{EVI}=2.5 \times \frac{(\mathrm{NIR}-\mathrm{Red})}{\left(\mathrm{NIR}+6 \times \mathrm{Red}-7.5 \times \mathrm{Blue}\right)}$) and the Moisture Stress Index ($\text{MSI}=\frac{\text{SWIR}_1}{\text{NIR}}$).
In this experiment, we compare the end-to-end feature extraction on the full Sentinel-2 band spectrum to single input channels, common combinations and domain-specific vegetation indices.
The channel combinations are: all Sentinel-2 input channels including RGB and all available near-infrared channels as in the previous experiment; S4 for the red, blue, and two near-infrared channels, which are used in the vegetation index calculation; NDVI, EVI and MSI as common domain-specific vegetation indices calculated through a linear combination of the S4 channels; RGB as only the three common color channels without near infrared information.
\begin{table}[t]
\centering
\caption{Performance comparison across different Sentinel-2 input band variations (all, S4, RGB) and vegetation indices (NDVI, EVI, MSI) for the ViT/2 model on the sugar beet yield prediction task.} 
\label{tab:input_channels}
\begin{tabular}{@{}cccc@{}}
\toprule
      \thead{Sentinel-2 \\ Bands} & \thead{Input \\ Channels} & \thead{Validation \\ RMSE} & \thead{Test \\ RMSE}  \\ \cmidrule(lr){1-4}
      All & 10 &   \textbf{17.12}                    &  \textbf{20.41}                      \\ %
      S4 & 4 &   17.66                   &  20.56                         \\ %
      NDVI,\,EVI,\,MSI & 3 & 17.43 & 20.58 \\ %
      RGB & 3 &   18.24                     &  21.10                   \\ %
      NDVI & 1 &   18.49                    &  22.13                     \\ %
      EVI & 1 &   18.49                    &  22.13                       \\ %
      MSI & 1 &   18.07                    &  21.32                       \\ 
      \bottomrule
\end{tabular}
\end{table}

As shown in table~\ref{tab:input_channels}, training on all Sentinel-2 bands outperforms any other hand-selected bands or combinations.
We hypothesize that, given the difficulty of the purely optical prediction task, removing channels limits the model's ability to learn nuanced differences across fields.
Interestingly, even domain-specific vegetation indices fall short of the learned band combinations from end-to-end training.
We hypothesize this is due to the linear nature of band combinations in the vegetation indices and the task-specific constants in the index computations that would require hyperparameter tuning.
The difference in the amount of parameters (281k-284k) and compute across the band configurations is negligible (292-295 GFLOPS) in our setting, as only the first patch embedding layer is affected.
Future work will include employing explainability techniques to understand what channel combinations were learned by the model through end-to-end training and how these relate to the domain-specific vegetation indices.
This knowledge could guide the development of even more specialized models. 

\subsection{How informative are the relative positions of field patches?}\label{sec:positional_encodings}
So far in our experiments, we have only relied on the optical data and excluded standard relative positional encodings for the patches.
Without positional encodings, the spatial relationships between the patches and the shapes of the fields are lost in the attention mechanism.
Since the yield per field is already normalized by the field size (t/ha), the field size should not affect the yield.
As the model could simply predict a yield per patch and aggregate the predictions for the entire field, we hypothesize that adding the relative positions of the patches does not influence performance. 

To test if the relative positional encodings for the patches benefit the model, we compare no positional encoding, a learned positional encoding and the fixed sinusoidal positional encoding~\cite{attentionisallyouneed}.
The results in table~\ref{tab:positional_encodings} indicate that adding positional encodings does not improve the model.
For the case of the learned encodings, the trainable parameter count almost doubles, which is difficult to optimize with the current size of the data set.
The fixed sinusoidal encodings do not add any meaningful computational overhead but also do not improve the model.
We hypothesize that the knowledge of relative patch positions is not necessary for our prediction task.
The padding may also play a role as the model does not need to first learn to separate fields from non-fields, for which relative positions could be beneficial.
Excluding relative positional encodings further prevents potential data leakage between the 2019 train set and the 2024 test set because information about the order of the field patches, and therefore the shapes of the fields, is lost.
We will omit positional encodings for the current experiments and explore possible explicit yield aggregation and positional encoding synergies in future work.

\begin{table}[t]
\centering
\caption{A comparison of different positional encoding strategies for the ViT/2 model on the sugar beet yield prediction task.}
\label{tab:positional_encodings}
\begin{tabular}{@{}cccc@{}}
\toprule
      \thead{Positional \\ Encodings}  & \thead{Validation \\ RMSE} & \thead{Test \\ RMSE} & \thead{Parameters \\ (million)} \\ \cmidrule(lr){1-4}
      None  &   \textbf{17.12}                    &  \textbf{20.41}    & 0.284                  \\ 
      Learned &   17.48                   &  21.00       & 0.546                  \\ 
      Sinusoidal & 17.44 & 20.44 & 0.284 \\ 
      \bottomrule
\end{tabular}
\end{table}

\subsection{How does model complexity affect task performance?}\label{sec:dim_heads}
After the task-specific model design, scaling up the model to trade-off compute and performance is a possible next step.
A common approach to scale the vision transformer is the embedding dimensionality for each patch and the amount of attention heads for the multi-head self-attention.
We explore the embedding dimensionality $d\in [64, 128, 192, 256, 320]$ and the number of attention heads $h\in [1, 2, 4, 8, 16]$.
We also tested multiple attention layers in sequence, but this lead to consistently worse performance with greatly increased compute requirements.

\begin{table}[t]
\centering
\caption{The influence of a higher embedding dimension $d$ and more attention heads $h$ on the ViT/2 model performance for the sugar beet yield prediction.}
\label{tab:vit_emb_heads}
\begin{subtable}{1\textwidth}
\begin{tabular}{@{}cc m{0.21cm} ccccc m{0.21cm} ccccc@{}}
      
\toprule
      \multirow{2}{*}{\thead{dimensions}} &  \multirow{2}{*}{\thead{GFLOPS}}& &\multicolumn{5}{c}{validation RMSE loss} & & \multicolumn{5}{c}{test RMSE loss}\\\cmidrule(lr){4-8}\cmidrule(lr){10-14}
       &  & $h =$ & 1 & \textcolor{gray}{2} & 4 & 8 & 16 & $h =$& 1 & \textcolor{gray}{2} & 4 & 8 & 16  \\ \midrule
      \textcolor{gray}{64} & \textcolor{gray}{295} & & 17.12 & \textcolor{gray}{17.16} & 17.10 & 17.02 & 17.10 & & 20.41 & \textcolor{gray}{20.33} & 20.21 & 20.29 & 20.01        \\% 
      128& 625 & & 17.02 & 17.06 & 17.06 & 17.01 & 17.13 & & 20.40 & 20.69 & \textbf{19.85} & 20.41 & 20.41
      \\ %
      192&  989 & & 17.06 & 17.20 & 17.00 & 17.07 & 17.03 & & 20.46 & 20.55 & 20.95 & 20.23 & 20.22 
      \\ %
      256 & 1387 & & 17.04 & 17.14 & 17.15 & 17.03 & \textbf{16.96} & & 20.35 & 20.55 & 20.12 & 20.20 & 19.96
      \\ %
      320 & 1820 & & 17.01 & 16.98 & 17.04 & 17.05 & 17.07  & & 20.63 & 20.35 & 20.24 & 20.06 & 20.21               
      \\ \bottomrule
\end{tabular}
\end{subtable}%
\end{table}
Table~\ref{tab:vit_emb_heads} shows the different ViT configurations with compute requirements and the respective performance. 
The configuration from the previous experiments is highlighted in gray and the best performing configuration is shown in bold.
Comparing different columns per row in table~\ref{tab:vit_emb_heads} shows that simply adding more attention heads does not improve performance.
Similarly, when comparing the rows in the same column, increasing the embedding dimensionality of each patch does not consistently lead to a lower error, but greatly increases the required compute.
We hypothesize that a higher embedding dimensionality, more attention heads and even more transformer layers would require a bigger data set to improve performance without overfitting on the fields in the training set.

\subsection{Learnings from the specialized vision transformer}\label{sec:summary}
\begin{figure}[t]
      \centering
      \includegraphics[width=0.9\linewidth]{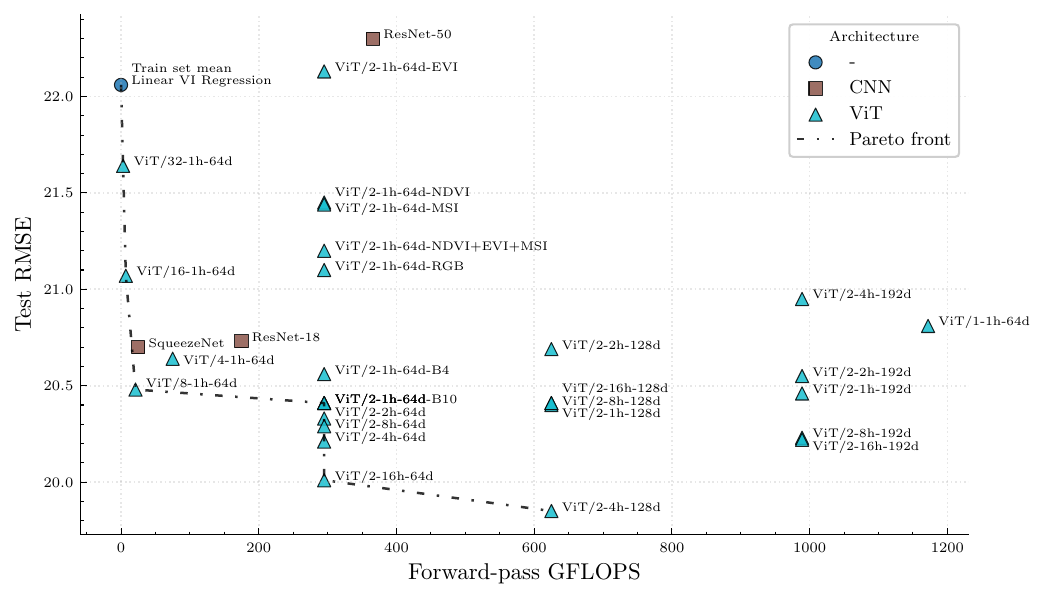}
      \caption{Compute requirements vs test set performance of multiple baselines and various vision transformer design choices on the sugar beet yield prediction task.}
      \label{fig:pareto}
\end{figure}

To put our findings into context, the pareto-style plot in figure~\ref{fig:pareto} shows the performance versus compute trade-off. 
The first baseline ViT model from table~\ref{tab:baselines} with a patch size of eight provides decent performance at very low compute.
The best performance is achieved be the ViT configuration with a small patch size of two with four attentions heads and a 128-dimensional embedding dimensionality.
The best compromise configuration is the ViT/2-16h-64d model with a small patch size of two, all spectral bands and additional attention heads.
For our goal to extract as much information as possible from purely optical Sentinel-2 data without extra conditioning information, we can conclude the following:
\begin{itemize}
      \item Incorporating domain knowledge into the model selection can lead to synergies. Our task requires fine-grained details between parts of the field, for which the transformer model operating over field patches performed the best.
      \item The receptive field of the model needs to be specific to the task. As the spatial resolution of our data at 10 m includes many plants, further compressing the information through larger patch sizes is detrimental to performance.
      \item End-to-end feature extraction should be explored alongside domain-specific input channel configurations. We have seen on our task that feature extraction on all bands outperforms band subsets or combinations. While this may have been due to the specific data setup and task, experiments on the full band spectrum can provide valuable insights for future model developments.
\end{itemize}

\section{Model Application}\label{sec:model_application}
For the final section, we put our model to use and predict the yield before the harvest through temporal encodings (section~\ref{sec:early_prediction}) and find underperforming fields through a ranking-based metric (section~\ref{sec:underperforming}).
We use the simple ViT/2-1h-64d model to keep compute requirements modest and to avoid overfitting when introducing additional temporal parameters.
We first expand the training to all time steps leading up to the harvest, not just the last one as in the previous experiments.
For the model training, we randomly sample one of the available temporal instances for each field per mini-batch and predict the harvest yield at the end of the growth cycle.
It is important to note that the model still only receives a single temporal instance for a prediction.
With the added temporal instances before the harvest to the training set and corresponding time encodings, the model can additionally learn how fields look like during the growth cycle given the yield after the harvest.
The satellite images for a field at different time steps can be vastly different depending on the growth cycle.
We evaluate multiple encoding strategies in section~\ref{sec:early_prediction} to assess the influence of time encodings on the ability of the model to learn time-aware features for the early prediction.

In our very restrictive setup, estimating the magnitude of yield purely based on a single optical satellite image is challenging by design ($R^2=0.137$ for the best model on the test data).
However, we hypothesize that the model may have learned to rank fields based on the harvest yield as it still outperforms simple baselines and does capture some variance of the test labels.
In this scenario, the correct magnitude is not important as long as the relative order of the yield predictions is correct.
We therefore use the model for the detection of low-yield fields early on in the growth cycle in section~\ref{sec:underperforming}. 

\subsection{Early yield prediction through time encodings}\label{sec:early_prediction} 
\begin{figure}[t]
      \centering
      \includegraphics[width=0.9\linewidth]{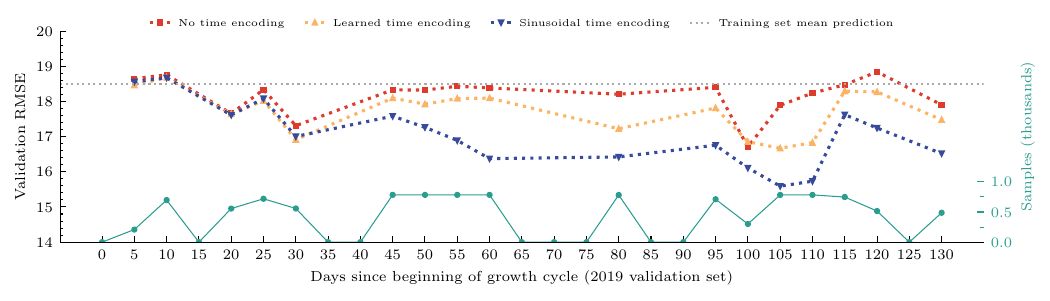}
      \includegraphics[width=0.9\linewidth]{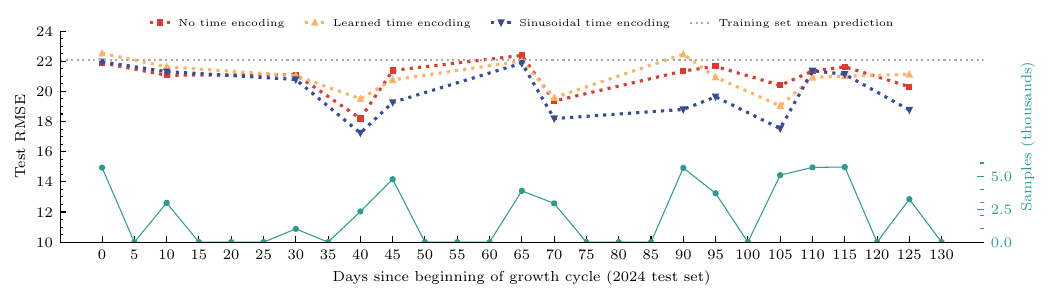} 
      \caption{Early detection performance for the single layer single head ViT/2 model and different time step encodings for the validation set (top) and the test set (bottom). The amount of valid instances is displayed in green for each data set.}
      \label{fig:early_detection}
\end{figure}

The results for different time encoding strategies are displayed in figure~\ref{fig:early_detection}.
We test no temporal encoding (red line), a learned temporal encoding (yellow line) and a fixed sinusoidal temporal encoding (blue line), which is added to each temporal instance before passing it to the transformer layer.
For the learned temporal encodings, almost no additional parameters are added to the model as there are only 26 distinct time steps in the growth cycle.
To evaluate the model, we iterate over the time steps and select all fields which have a temporal instance at that time step.
We only show predictions for time steps with a large enough sample size, where at least 20\% of the fields have a valid temporal instance.

The results in figure~\ref{fig:early_detection} show the performance of different temporal encoding strategies over different time steps in the growth cycle on the top (red, blue, yellow line). 
The corresponding sample sizes are shown on the bottom (green line).
We can see that the sinusoidal time encoding consistently improves performance.
Despite filtering time steps with low sample size, the results still show high variance across the growth cycle, especially for the test set.
We hypothesize this is due to an inconsistent amount of temporal instances per time step during training.
Furthermore, as the test set was recorded in 2024, whereas the training data was from 2019, the dates do not perfectly align and differences in weather conditions lead to a different distribution of valid temporal instances across the time steps.
The "day 0" also differs between years as the growth seasons did not start at exactly the same time.
However, the low error very early in the growth cycle (day 30 on the validation set and day 40 on the test set) indicates that early prediction is possible and interesting for future work.

\subsection{Early detection of underperforming fields}\label{sec:underperforming}
For the final experiment, we will only use the harvest yield as a proxy metric to find underperforming fields through our model.
First, we define \textit{underperforming fields} as the fields in the lower quartile of harvest yield.
As the harvest yield is given in tons per hectare, so already normalized by the area of each field, the differences in yield should be independent of the field size.

\begin{figure}[t]
      \centering
      \includegraphics[width=1\linewidth]{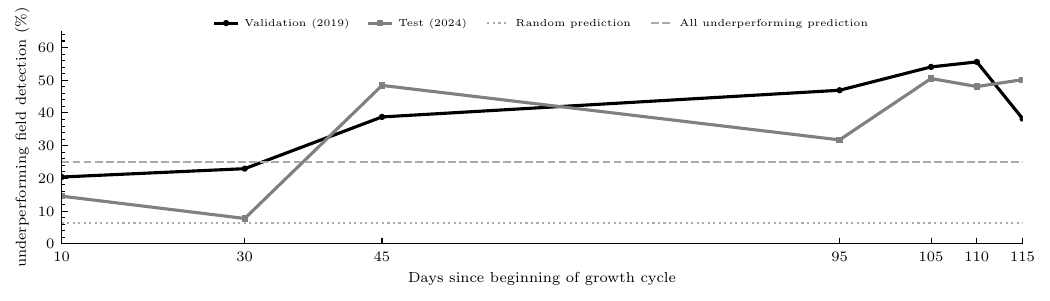} 
      \caption{Detection percentage of the fields in the lowest quartile of harvest yield across time steps by the ViT/2-1h-64d model with sinusoidal time encodings.}
      \label{fig:underperformers}
\end{figure}

In our experiment, we predict the yield for all available temporal instances for the validation and test set at multiple time steps, similarly as in the previous section~\ref{sec:early_prediction}.
We then rank the fields by harvest yield, compare the lowest quartile of predicted fields to the underperforming fields based on the ground-truth labels, and report the percentage of fields in both sets.
As 25\% of fields are underperforming based on this definition and the ground-truth harvest yield, a random prediction of underperforming fields should identify a quarter of the real underperforming fields, or 6.25\%, indicated by the dotted line in figure~\ref{fig:underperformers}.
Classifying all fields as underperforming would lead to a 25\% detection rate, shown as a dashed line in figure~\ref{fig:underperformers}.
Note that we only select the time steps where there are a lot of fields with a valid temporal instance for both validation and test sets, such that the quantity of predictions is comparable to the amount of ground-truth underperforming fields.
However, not every underperforming field necessarily has a valid temporal instance at each selected time step, which can reduce the observed detection rate.

Despite this strong data limitation, the results in figure~\ref{fig:underperformers} show that almost 50\% of underperforming fields were already found one third of the way into the growth cycle.
In practice, this could be used to alert farmers that their field is underperforming way before the harvest when interventions are still possible. 
Surprisingly, this was possible through purely optical data.
Integrating more sources of information and further developing the approach will be subject to future work.
\section{Conclusion}
Applying machine learning methods to real-world tasks is non-trivial. 
Practitioners must balance generic deep neural network architectures against domain-specific pre-processing steps, such as band selection, that can both constrain and enhance model performance.
This paper should provide a real-world example of this trade-off.
We showed that modeling choices driven by domain knowledge can lead to synergistic effects and even surprising results.
Our experiments show that small input patches lead to consistently better performance, which contrasts with most previous, predominantly convolution-based work that relies on progressively larger receptive fields. 
We also find that using all Sentinel-2 spectral bands, without extensive pre-processing or custom band combinations, is beneficial - another uncommon design choice in the field.
The goal of the paper was not to provide a new prediction system that outperforms previous approaches, but instead highlight the marginal improvements of key design choices as insights for future practitioners.
Despite the limited direct practical use of our absolute yield predictions, the study allowed us to systematically question critical design choices and demonstrate performance improvements grounded in domain knowledge. 
An important promising result of our study is the ability for the same models to detect underperforming fields well before the harvest when combined with an appropriate training setup and a ranking-based detection.
Our study is limited by the small dataset size, the focus on specific years, a single geographical region, one crop type, and one satellite mission.
Future work will verify these findings in broader settings; a more detailed discussion of the limitations is provided at the end of section~\ref{sec:methods}. 

\bibliographystyle{splncs04}
\bibliography{mybibliography}

\end{document}